# A Novel Grip Force Measurement Concept for Tactile Stimulation Mechanisms – Design, Validation, and User Study

Guy Bitton, *Member, IEEE,* Ilana Nisky, *Senior Member, IEEE*, David Zarrouk

*Abstract*— We developed a new grip force measurement concept that allows for embedding tactile stimulation mechanisms in a gripper. This concept is based on a single force sensor to measure the force applied on each side of the gripper, and it substantially reduces artifacts of force measurement caused by tactor motion. To test the feasibility of this new concept, we built a device that measures control of grip force in response to a tactile stimulation from a moving tactor. First, we used a custom designed testing setup with a second force sensor to calibrate our device over a range of 0 to 20 N without movement of the tactors. Second, we tested the effect of tactor movement on the measured grip force and measured artifacts of 1% of the measured force. Third, we demonstrated that during the application of dynamically changing grip forces, the average errors were 2.9% and 3.7% for the left and right sides of the gripper, respectively. Finally, we conducted a user study and found that in response to tactor movement, participants increased their grip force, and that the increase was larger for a smaller target force and depended on the amount of tactile stimulation.

## I. INTRODUCTION

During daily activities, we manipulate objects and secure them from slippage by means of stabilizing grip forces. For example, when we push a pen against a paper we apply a grip force using our fingers and adjust it to prevent the pen from slipping. Aside from natural interaction with objects, grip force on a manipulation mechanism is also important in non-natural interactions, such as teleoperation [1] or virtual reality with haptics [2]. At the master interface, it can be an important control signal for manipulation of objects with the remote tool [3].

In these applications, feedback about the interaction with the remote or virtual objects is provided at the manipulator interface. In recent years, there is an increasing popularity of the design of tactile devices for such feedback [2]–[22]

The study is supported by the Binational United-States Israel Science Foundation (grant no. 2016850), by the Israeli Science Foundation (grant 823/15), by the Ministry of Science and Technology via the Israel-Italy virtual lab on Artificial Somatosensation for Humans and Humanoids, and by the Helmsley Charitable Trust through the Agricultural, Biological and Cognitive Robotics Initiative of Ben-Gurion University of Negev, Israel.

G. Bitton and D. Zarrouk are with the Department of Mechanical Engineering, Ben-Gurion University of the Negev, Beer-Sheva, Israel (e-mail: guybit@post.bgu.ac.il).

I. Nisky is with the Department of Biomedical Engineering and Zlotowski Center for Neuroscience, Ben-Gurion University of the Negev, Beer-Sheva, Israel (e-mail: nisky@bgu.ac.il).

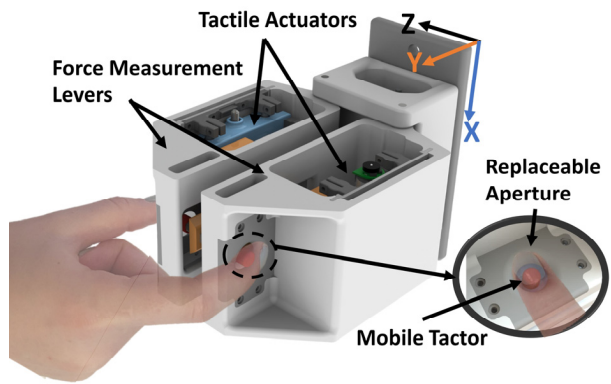

Fig. 1. A general overview of the grip force measurement device. The user holds the device by placing the finger and thumb on the apertures from opposite sides of two identical levers. The levers are fitted with a tactile stimulation mechanism that moves a rubber tactor against the skin (inset shows zoomed in top view of a semi-transparent index finger on the aperture). The device measures each of the forces acting on the two levers using a single force sensor (not visible).

However, to date, the design of most of the devices focuses largely on providing compelling tactile sensation without accurate measurement of the motor response and specifically the adjustment of grip force. By understanding the effect that tactile stimuli have on the adjustment of grip force, we can eventually design better manipulation mechanisms and controllers for teleoperation.

Recent studies incorporated a grip force measurement element into their tactile stimulation devices and tested the effect that skin-stretch stimuli have on grip force control [5], [6],[13],[15]. However, those studies recorded only part of the grip force. As a result, the measured grip force was scaled down and may have suffered from stimulation artifacts [15].

In this study, we present a new versatile grip force measurement concept for tactile devices, capable of accurately measuring grip force from each side of the gripper separately using a single force sensor and minimizing artifacts. We implemented this concept on a static platform and present its design followed by the design and control of a new 2-DoF skin-stretch device that we integrate into it (Fig. 1). Furthermore, we present an analysis of the measured forces and the measurement artifacts. To validate our force measurement concept, we calibrated the device and quantified the artifacts using an external force measurement system. Finally, we demonstrated the use of our device in a user study in which participants were instructed to maintain a constant grip force and tested how they responded to skin-stretch stimulation of the finger pad by adjustment of grip force.

## II. BACKGROUND

Our novel device combines a grip force measurement mechanism with tactile stimulation by means of a tactor movement that stretches the skin of the finger pads (Fig. 1). In this section, we provide background on grip force control, tactile stimulation, and the connection between them.

### A. Grip Force Control

Grip force is the force that we apply perpendicularly to objects held in precision grip to prevent their slippage. Human control of grip force consists of feedforward and feedback components [23],[24]. Feedforward control of grip force is composed of a nonspecific baseline and modulation in anticipation of the load forces that will be applied by the object. The baseline component is increased with uncertainty about the load force [2] while the anticipation of load force is based on an internal model of the object dynamics, which is updated during repeated interactions with the object [23]–[31]. The feedback component adds an additional modulation in grip force in reaction to an indication of slippage or a sudden load force by cutaneous mechanoreceptors [23],[25], [29],[32]–[35].

Most of the studies that aimed at understanding the contribution of tactile information to grip force control involved anesthetized [36] or impaired tactile sensing subjects [27]. Fewer studies investigated the contribution of tactile information to grip force control in the intact motor system. Quek et al. [5] incorporated a grip force measurement element into their 3-DoF skin-stretch device to test the effect of skin-stretch on the average grip force. In [37], a similar device was used to test the effect on the predictive and reactive components of grip force control. In both studies [5],[15], the skin stretch mechanism consisted of an aperture and a mobile tactor that stretched the skin relative to the aperture, and both devices measured only the part of the grip force which is acting on the aperture, omitting the component acting on the tactor. This partial measurement limited the interpretation of the experimental data because it was not clear whether changes in the measured forces were fully or partially caused by artifacts [37]. The grip force measurement concept that we present here is designed to minimize artifacts and allow for a reliable interpretation of changes in grip force in response to tactile stimulation.

Moreover, the studies of grip force control mostly addressed the total or average grip force coming from both sides of the grip [1],[6],[15],[26],[29],[36]. Indeed, when a held static object is not grounded, the grip forces on both sides must be equal to prevent movement. However, a simple force analysis can show that it is not the case when the object is grounded, moves dynamically or when the grip is not horizontal. Therefore, several studies measured the force applied by each of the gripping fingers separately to shed light on the control of grip force during object manipulation [38]–[47]. For example, Edin et al [47] used separate force sensors for the index finger and the thumb and found that the force applied by each digit is controlled independently during precision lifting. Metzger et al [42] incorporated two force sensor into their robot to asses hand functions. Other studies used separate sensors to investigate abnormalities in the control of grip in pathological conditions, such as in Parkinson's disease [48],[49] or stroke [50].

The measurement of grip force in each of the fingers presents design challenges to the incorporation of accurate force sensors in a gripper due to space limitations [38]–[45]. This challenge is exacerbated if one wishes to accurately measure the grip force during tactile stimulation [5],[6],[13], [15]. In some cases, the use of multiple sensors that are in direct contact with the fingers is necessary, such as in unconstrained grasping [43] or in the measurement of digit placement [44]. Nevertheless, in other cases, reducing the number of sensors and positioning them away from the digits can be beneficial. Our novel grip force sensing concept uses a single 6-axis force sensor that is located away from the grip site to measure the grip forces, and hence solves both design challenges simultaneously while reducing the number of expensive force sensors.

### B. Tactile Stimulation Devices

Traditional haptic devices are kinesthetic, i.e., devices that can apply forces and torques on the user. These devices provide a complete haptic experience since they both apply forces and stimulate the skin [51]. Despite their many advantages, they bring a challenge of instability [52],[53], which calls for alternative ways to provide haptic feedback by means of tactile stimulation.

The simplest form of tactile stimulation is vibrotactile feedback which is already used broadly in mobile phones and other electronics to convey binary information, i.e., an indication that something occurred. The dimensionality of the conveyed information can be increased by including arrays of vibration actuators [54] or by altering the frequency and amplitude of the vibration [55].
Another recently developed tactile feedback approach is skin deformation. The deformation can be in the normal [18] or in the tangential direction [22] causing indentation or stretch of the skin, respectively. Skin stretch devices can produce deformation that resembles the deformation occurring in natural interactions with objects, and can convey object properties such as stiffness [4], curvature [21] or friction [14] . It can also convey artificial information for sensory substitution in prosthetics [12] or teleoperation [7],[22]. Skin stretch devices can be distinguished by the location on the skin being stretched. Some common locations for using skin stretch devices are the palm [9], forearm [11], upper arm [56] and finger pad [16]. The latter has the largest density of mechanoreceptors compared to other locations on the skin [57] and was therefore the location of choice for the design of the skin-stretch device in this study.

Additionally, an important characteristic of skin stretch devices is the attachment of the device to the body. A common attachment method for the finger-pad is that of a thimble [16]. Although thimble-based attachment can be effective in lab use [58], it may damage the natural feeling of the interaction with objects and can be inconvenient to use with some portable devices. Another attachment method is that of an aperture [19]. The pressure against the aperture prevents tangential movements of the finger and isolates the skin from skin deformation caused by other interaction forces with the device while providing a more natural interaction with the skin. In the design of our device, we chose to use aperture-based attachment to provide an interaction with the gripper that resembles that of holding a portable object.

Finally, skin-stretch devices are also differentiated by their number of degrees of freedom (DoFs). 1-DoF devices [22], [56] stretch the skin in a single direction and can convey information about magnitude using the amount of stretch [4]. 2-DoF devices can move and convey directional information [17] while 3-DoF devices also cause normal deformation of the skin [5],[8]. Additional degrees of freedom can be implemented using rotating elements that cause torsional skin deformation or by applying skin stretch in opposite directions to create a torsional sensation [20],[6]. We chose to implement a 2-DoF actuator to allow for investigation of skin-stretch in all the directions in the contact plane. However, we test only one direction in our user study.

III. DESIGN, CALIBRATION, AND VALIDATION

A. Design

Our device (Fig. 1) incorporates two main components: a force measurement system and a stimulation mechanism. The force measurement system is composed of two levers and a force sensor to measure the grip forces (Fig. 2(a)), and the stimulation mechanism (Fig. 2 (b)) is mounted on the levers and moves a tactor to stretch the skin of the finger pad. The tactor is a rubber Lenovo® TrackPoint® whose diameter is 2 mm (identical to the one used in previous devices [22]). As the user holds the levers and applies grip force against the apertures, the movement of the tactor stretches the skin (Fig. 1). The device is primarily manufactured using 3D printing.

*1) Force Measurement System*

The force measurement system is composed of two identical levers attached to a base which holds a 6-DoF ATI® NANO25® force sensor, through a rotational axis. The grip forces applied by the user are rotated and amplified using the rotational joint. The sum of the amplified grip forces acting on both levers is measured by the 6-DoF force sensor. The design distance between the center of the force sensor to the contact points of the lever (marked as d) is 7.32 mm. As a result, each of the grip forces applies a torque on the force sensor in the $y$ direction which is proportional to the force. As the first lever produces a positive torque, the second lever produces a negative torque, and hence, the measured torque is proportional to the difference between the two grip forces. Using the sum and difference of the grip forces we can evaluate the value of each of them separately (see section B).

*2) Tactile Stimulation Mechanism*

A tactile stimulation mechanism, such as the skin-stretch mechanism shown in Fig. 1 and Fig. 2, can be mounted into each one of the levers. Here, we present the design of an actuation mechanism for the tactors composed of two identical uncoupled linear actuators. The linear actuators of each mechanism are orthogonal; the $y$ actuator is mounted on top of the link of the $x$ actuator such that they can produce motion in the $x$ and $y$ directions (directions are marked in Fig. 2). Each actuator is comprised of a 12 mm Polulu® micro DC motor with a 1:51.45 gear box producing a maximum torque of 84.37 Nmm and a nominal speed of 590 RPM. The motors rotate 4 mm leadscrews whose pitch $p$ is 0.7 mm. As a result, the linear motor advances its link by 0.0136 mm per motor revolution which ensures high positional accuracy. Given the input torque of the motors, the diameter of the leadscrew, and the coefficient of friction of steel over steel (0.17) [59], the linear actuator can produce an estimated thrust force of 181.8 N [59]. A small arm, which we refer to as the end link, is rigidly connected to the second link. The tactor, a Lenovo® TrackPoint® Cap which is in touch with the user's finger pad, is attached to the end link through a 5 mm screw that allows the adjustment of its height and therefore indentation into the skin of the user. The tactor can be easily replaced if different contact characteristics are required.

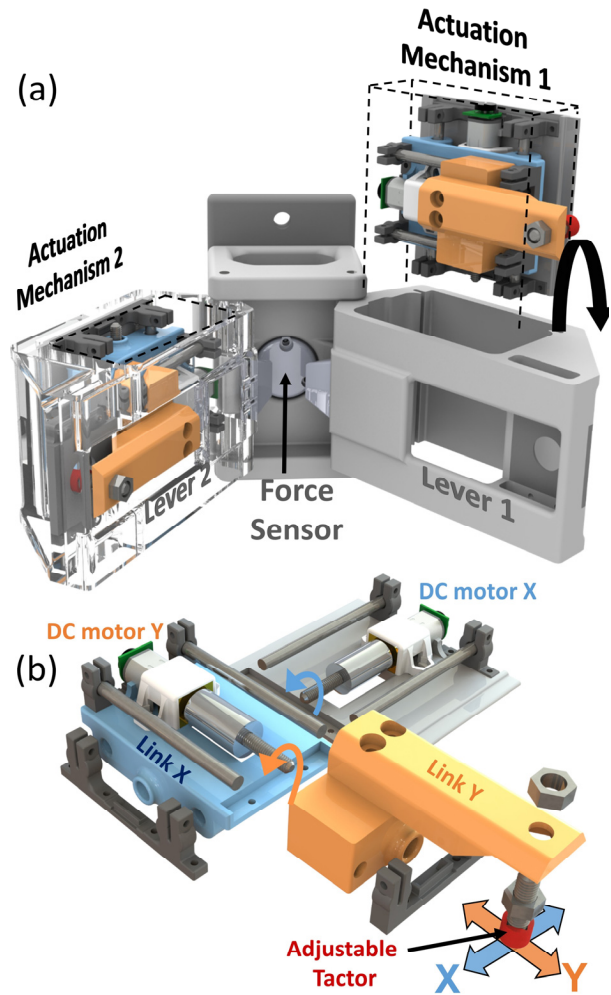

Fig. 2. The main components of the device. (a) The back base with the force sensor, the two levers and the actuators of the tactors integrated into each lever. (b) An exploded view of the actuator of the tactor composed of two identical and uncoupled linear actuators ($x$ and $y$ directions).

*3) Control and Electronics*

The motor position is controlled using closed loop Min-max control on the duty cycle at a rate of 100 Hertz. Each of the motors is controlled separately. We used an Arduino® Uno® controller to control the motors using an 8 volts battery power source and a BasicMicro® RoboClaw® motor driver to amplify the input power and current. Magnetic encoders were instrumented on the main axis of each of the motors to produce 617 counts per output revolution. The closed loop error is smaller than 60 counts which is equivalent to 0.068 mm, for the given gear ratio and pitch.

To calibrate the zero position of each actuator, we used the current reading from the motor driver. The actuator first moves the link towards one direction until an increase in current is identified signifying that the link is blocked and reached its range limit. The actuator then moves the link to the middle of the predefined range and sets that position as zero. This ensures that the zero position is the same on different uses of the device. We used MATLAB® to communicate with the Arduino controller through a serial port to set the desired position.

*B. Force Analysis*

A force analysis is shown in Fig. 3. Each finger applies a grip force $F_{Grip}$ on the device, that is divided into $F_T$ and $F_A$ that are applied on the aperture and tactor, respectively, and are distributed over their contact area as demonstrated in Fig. 3. As the tactor moves, a friction force between the tactor and the contact point with the finger $F_{Tf}$ is generated. Moreover, a movement in the $y$ direction causes a change in the point of action of $F_T$ and therefore a small change in the calculation of the torque. Using the torque equilibrium around the joints (1 or 2), the force acting on the force sensor as a function of the grip force is:

$$F_M = \frac{L_G}{L_M}\left(F_{Grip} + \frac{L_A}{L_G}F_{Tf}\sin\theta + \frac{\Delta y}{L_G}\cdot F_T\right), \quad (1)$$

where $L_G$, $L_M$, and $L_A$ are the geometric dimensions of the levers, as shown in Fig. 3, and $\theta$ is the angle of the tactor movement with the $x$ axis. Since we are interested in $F_{Grip}$, we define the second and third terms in the parentheses as unwanted measurement artifacts, marked together $A_T$. Furthermore, we note that due to manufacturing tolerances some geometrical differences between the two levers might arise. Thus, considering both sides of the grip, we have

$$F_M = \frac{L_G^{(1)}}{L_M^{(1)}}\left(F_{Grip}^{(1)} + A_T^{(1)}\right) + \frac{L_G^{(2)}}{L_M^{(2)}}\left(F_{Grip}^{(2)} + A_T^{(2)}\right). \quad (2)$$

The grip forces also generate a torque $T_M$ given by:

$$T_M = \frac{L_G^{(1)}\cdot d_1}{L_M^{(1)}}\left(F_{Grip}^{(1)} + A_T^{(1)}\right) - \frac{L_G^{(2)}\cdot d_2}{L_M^{(2)}}\left(F_{Grip}^{(2)} + A_T^{(2)}\right). \quad (3)$$

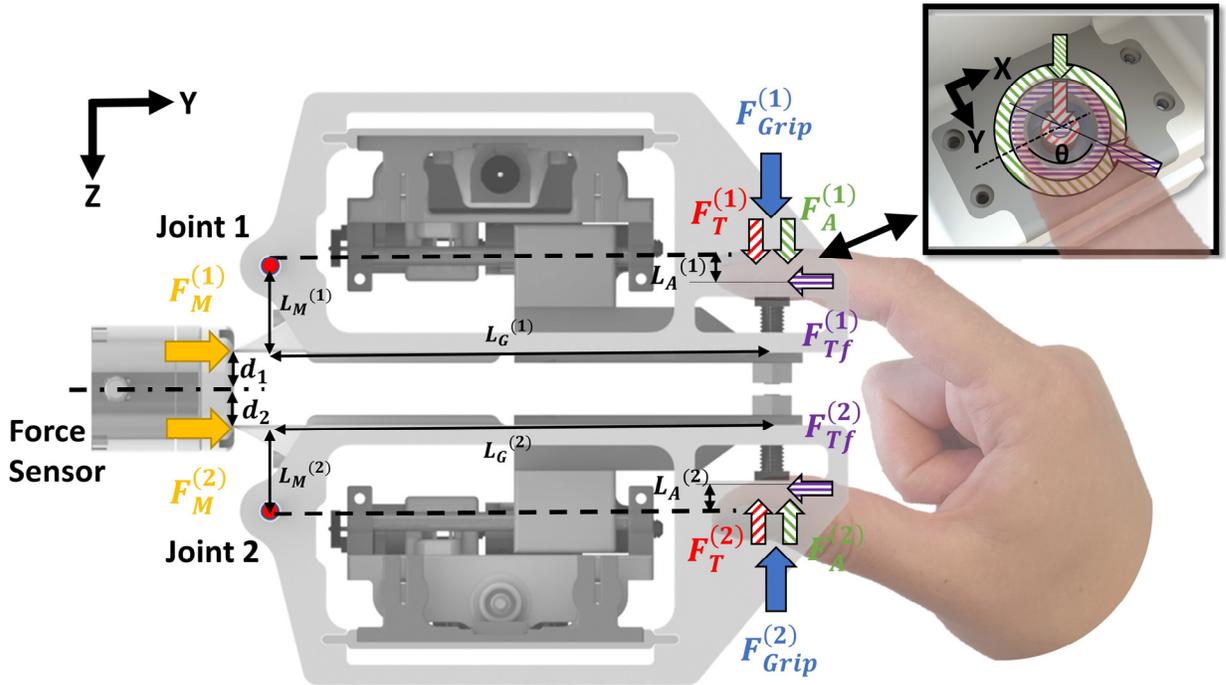

Fig. 3. Force analysis diagram. The force $F_M$ measured by the force sensor is the sum of the grip forces amplified by a factor of $L_G/L_M$, with an artifact caused by friction with the tactor and by a change in the point of action of $F_T$. The measured torque is proportional to the difference between the forces. The inset shows the contact areas of the forces $F_T$ and $F_A$ constructing $F_{Grip}$, and of the friction force $F_{Tf}$.

Solving (2) and (3) for $F_{Grip}^{(1)}$ and $F_{Grip}^{(2)}$, we have:

$$F_{Grip}^{(1)} = \frac{L_M^{(1)} \cdot d_2}{L_G^{(1)} \cdot (d_1 + d_2)} F_M - \frac{L_M^{(1)}}{L_G^{(1)} \cdot (d_1 + d_2)} \cdot T_M - A_T^{(1)}, \quad (4)$$

$$F_{Grip}^{(2)} = \frac{L_M^{(2)} \cdot d_1}{L_G^{(2)} \cdot (d_1 + d_2)} F_M + \frac{L_M^{(2)}}{L_G^{(2)} \cdot (d_1 + d_2)} \cdot T_M - A_T^{(2)}. \quad (5)$$

We mark:

$$\alpha_1 = \frac{L_M^{(1)} \cdot d_2}{L_G^{(1)} \cdot (d_1 + d_2)}, \quad \alpha_2 = \frac{L_M^{(2)} \cdot d_1}{L_G^{(2)} \cdot (d_1 + d_2)}, \quad (6)$$

and

$$\beta_1 = \frac{L_M^{(1)}}{L_G^{(1)} \cdot (d_1 + d_2)}, \quad \beta_2 = \frac{L_M^{(2)}}{L_G^{(2)} \cdot (d_1 + d_2)}. \quad (7)$$

Taking (6) and (7) and substituting into (4) and (5), we can calculate the grip force of each side separately using the measurements of $T_M$ and $F_M$ from one force sensor using

$$F_{Grip}^{(1)} = \alpha_1 \cdot F_M + \beta_1 \cdot T_M - A_T^{(1)}, \quad (8)$$

and

$$F_{Grip}^{(2)} = \alpha_2 \cdot F_M - \beta_2 \cdot T_M - A_T^{(2)}. \quad (9)$$

$A_T^{(1)}$ and $A_T^{(2)}$ are unknown, thus they contribute to the error but are not included in the calculation of $F_{Grip}^{(1)}$ and $F_{Grip}^{(2)}$, as shown in Section C. For cases of no tactor movement in the $y$ direction, the theoretical artifacts in (1) - (9) are all zero. Note that $F_{Grip}^{(1)}$ is not equal to $F_{Grip}^{(2)}$, and to provide a measurement that is consistent with other studies of the control of grip force, we define the mean grip force as

$$F_{mean} = \frac{F_{Grip}^{(1)} + F_{Grip}^{(2)}}{2}. \quad (10)$$

This force is used in the user study (Section IV) as the target force in the experiments.

### C. Calibration & Validation
#### 1) Calibration of Force Measurement

To calibrate our force measurement system, we designed a 3D printed stand instrumented with a force sensor (ATI® nano17®) to measure the applied grip force, as depicted in Fig. 4. An adjustable link, which can move vertically using a lead screw, is connected to the sensor. To simulate the force interaction of a finger and distribute the force between the tactor and the aperture, the tip of the link in contact with the lever is made of DragonSkin® NV10 silicon rubber.

During the experiments, we lowered the adjustable link to apply forces ranging from 0 to 20 N on the tactor and aperture while measuring the force and torque acting on the base sensor. We used linear regression in accordance with (2) and (3) to find the ratio of the distances $L_G / L_M$ and their product with the distance $d$ for both levers. The linear fit had $R^2 > 0.99$ for all cases as presented in Fig. 4. Using these results, we found $d_1 = 7.17$ mm and $d_2 = 5.98$ mm (which is consistent with visual measurements). Although the geometrical measurements of both levers are designed to be symmetrical, manufacturing and assembly tolerance caused the large difference between $d_1$ and $d_2$, which is accounted for in our calculations. We used (6) and (7) to calculate $\alpha_1$, $\alpha_2$, $\beta_1$ and $\beta_2$. The values of all of the above parameters are summarized in Table I.

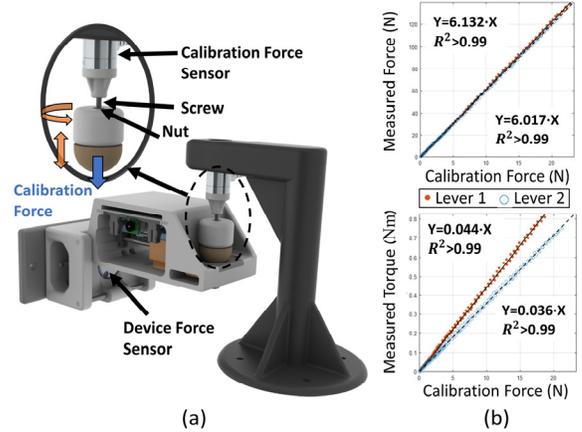

Fig. 4. (a) The experimental setup used to calibrate the device. During the experiment, the adjustable link was lowered to apply forces ranging from 0 to 20 N over the tactor area while measuring the force and torque acting on the device's force sensor. (b) Results of the calibration with fitted linear lines in accordance with (2) and (3). The results were used to find the coefficients $\alpha$ and $\beta$ of (8).

TABLE I.

DEVICE'S PARAMETERS CALCULATED IN THE CALIBRATION PROCESS

| Parameter | Value | |
|---|---|---|
| | Lever 1 | Lever 2 |
| $L_G / L_M$ | 6.132 | 6.017 |
| $(L_G / L_M) \cdot d$ | 0.044 m | 0.036 m |
| $d$ | $7.17 \cdot 10^{-3}$ m | $5.98 \cdot 10^{-3}$ m |
| $\alpha$ | 0.074 | 0.091 |
| $\beta$ | 12.39 m | 12.63 m |

#### 2) Characterization of Artifacts

In addition to the measurement artifacts described in (2) - (7), another cause for measurement artifacts is a non-uniform distribution of grip force over the tactor and aperture during tactor movement. To measure the actual artifact, we used the calibration system described in Fig. 4 to apply a force of 15 N on the tactor and aperture of one lever. We then moved the tactor back and forth by a distance of 1.5 mm (Fig. 5 (b)), while measuring the force using the calibration system's force sensor and the device's force sensor. We repeated this in both the $x$ and $y$ directions (Fig. 5(a)). Note that we chose the values of 15 N and 1.5 mm to account for the most extreme cases that the haptic device was designed for. As we

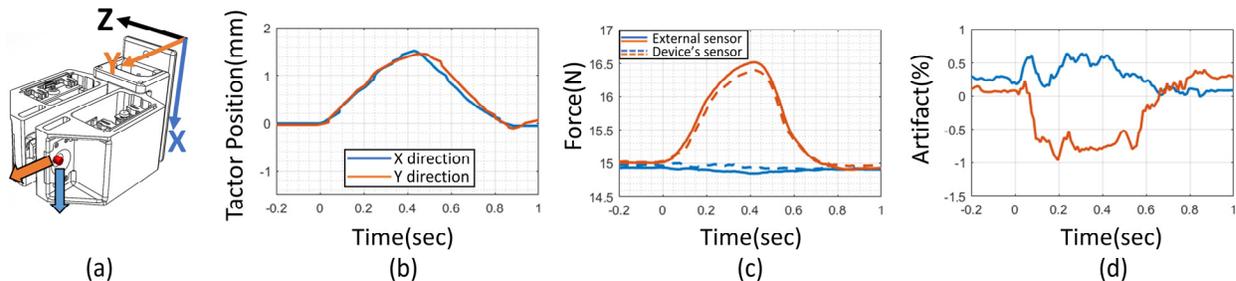

Fig. 5. Characterization of artifacts for 15 N grip force and 1.5 mm tactor movement. Using our calibration system, we applied a 15 N force to the device and recorded the grip forces using the device's sensor and the external sensor during a tactor movement of 1.5 mm and back in both the *x* and *y* directions. (a) Coordinate system of the device. (b) The tactor movement trajectory. (c) Change in force during tactor movement. (d) The calculated measurement artifact during tactor movement.

are interested in the relative artifact error due to tactor movement $A_{TM}$, we define

$$A_{TM} = \frac{F_{External} - F_{Device}}{F_{External}}, \quad (11)$$

where $F_{External}$ is the force measured by the external force sensor and $F_{Device}$ is the force calculated from the measurement of the device.

Fig. 5(b) presents the tactor position trajectory during the movement in the *x* and *y* in the *x* and *y* directions, Fig. 5(c) presents a comparison between the forces measured by the external sensor and the haptic device's sensor during that movement and Fig. 5(d) presents the relative artifact $A_{TM}$ using (11). The maximum measured relative artifact $A_{TM}$ is 0.65 % and 1.03 % of the applied external force for tactor movement in the *x* and *y* directions, respectively.

*3) Separate Measurement of Forces Applied Simultaneously on Both Sides of the Gripper*

To evaluate the accuracy of the device in simultaneously measuring forces from both levers, i.e. in this case simultaneously measuring the force applied by the finger and thumb, we conducted a test using 2 external ATI® nano17® force sensors attached to each of the levers (Fig. 6). Each of the sensors was positioned between the aperture and the finger or thumb to directly measure the forces.

We applied a series of three different grips: a grip with a larger force applied by the finger (right), followed by a grip with a larger force applied by the thumb (left) and a third grip in which we attempted to apply an equal force with both finger and thumb. The results are shown in Fig. 6. The error (defined as the difference between the forces measured by the external force sensors to the force measured by the haptic device) averaged for $F_{Grip}>2$ N over the three grip tests is 2.93%, 3.71%, and 2.67%, for the right, left and mean force, respectively.

IV. USER STUDY

To demonstrate the use of our device with human participants, we performed a user study that included one experiment similar to the constant grip force experiment performed by Farajian et al. [15] (presented in appendix 1 of their study as "reactive experiment 2").

A. *Experimental Setup*

Eleven Subjects (5 males and 6 females, ages 24-40) participated in this study and all reported to be right-handed. One participant (female) was excluded from the analysis since she reported to have not followed experimental instructions properly. The study was approved by the Human Subjects Research Committee of Ben Gurion University of the Negev, Be'er-Sheva, Israel. All participants signed an informed consent form.

As depicted in Fig. 7, participants were instructed to sit comfortably in front of a computer screen and position their elbow on the table such that they can easily reach the device which was rigidly attached to the table. A barrier prevented them from seeing the device along the experiment. The participants were instructed to hold the device using their finger and thumb while keeping them horizontal. To simplify the experiment, we actuated the right tactor only such that the skin-stretch stimuli were only applied to the finger. In addition, the tactor was moved in the vertical direction only (*x* direction), i.e. the actuation device was used as a 1-DoF skin stretch device. All participants wore noise-cancelling headphones to avoid sound cues.

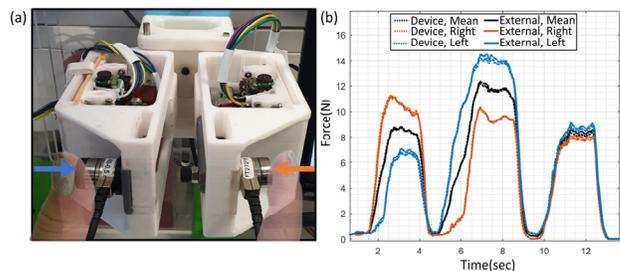

Fig. 6. Demonstration of the accuracy of the device simultaneously recording forces from both sides of the grip. (a) The experimental setting. Forces were recorded with the device's force sensor and external force sensors simultaneously, while applying forces on both sides of the device in a series of three types of grips: right force>left force, left force>right force, and a third one trying to keep the force on the right and left as equal as possible. (b) The measured and calculated forces. Dotted lines represent the forces calculated using (8), while the solid lines represent the external force sensors measurements. The three graphs correspond to the right(orange) and left(blue) sides separately, and the mean of both sides(black).

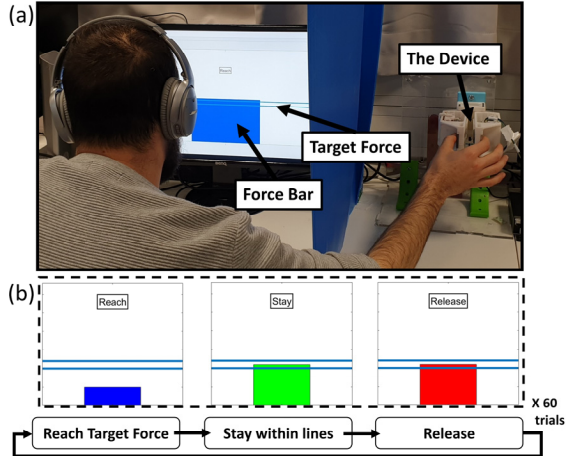

Fig. 7. The experimental setting. (a) The user holds the device with finger and thumb as demonstrated. A bar represents the mean grip force that is to be maintained between the horizontal lines designating target force). (b) The experimental protocol. Users had to reach a target force. After a randomized waiting time at the target level, a skin-stretch stimulus was applied. Trials were performed with three tactor displacements and two target force level, giving a total of 60 trials, 10 for each condition.

## B. Experimental Procedure

Each participant performed a total of 70 trials. To ensure that the initial response to the skin-stretch stimulus is not caused by unfamiliarity with the stimulus and that the participants follow the instructions properly, the first 10 trials were dedicated to training.

Each trial consisted of 4 phases. First, the ramp-up phase, in which the participants had to enhance their grip to bring the force bar in between the target lines. Second, the stable grip phase, in which the participants were requested to keep the force bar within the lines. Third, the stimulus phase, which starts after a randomized and predetermined time from the beginning of the stable grip phase (between 1 to 4 seconds). In this phase, the skin stretch stimulus was applied to the participants' finger pad while the participants were requested to maintain the force bar within the lines. Fourth and last, a waiting phase of 3 seconds after the beginning of the stimulus. At the end of the fourth phase the participants were asked to release the grip.

In each trial, the tactor displacement was either 0.5 mm, 1 mm, or 1.5 mm in the *x* direction and the target force was either 5 N or 7.5 N (6 distinct trials). The 60 trials of the experiment were constructed of 10 different blocks of 6 trials each. Within each block, all 6 combinations of tactor displacement and target force were randomized and predetermined. The trials were similar for all participants. Along the experiments, our device recorded grip forces and tactor position as a function of the time.

To analyze the change in grip force during tactor movement, we observed several trials of each of the subjects. We quantified the effect of tactor displacement on the change in grip force by defining the force difference $\Delta_{PS}$ as the difference between the peak force during tactor movement and the force recorded at the time the tactor starts to move, as marked in Fig. 8. Furthermore, we averaged the grip force for all 6 conditions (2 levels of target force, 3 levels of tactor displacement) across all 10 subjects, with t=0 sec defined as the beginning of tactor movement. Finally, we compared the average $\Delta_{PS}$ data for the different conditions. To test the statistical significance of the observed differences, we conducted a two-way repeated measures ANOVA on the $\Delta_{PS}$ data, and performed planned comparisons with Holm correction within each one of the two levels of target force. For these comparisons, we used pooled variance and the degrees of freedom associated with this pooled variance [60]. Results are considered significant at $\alpha=0.05$.

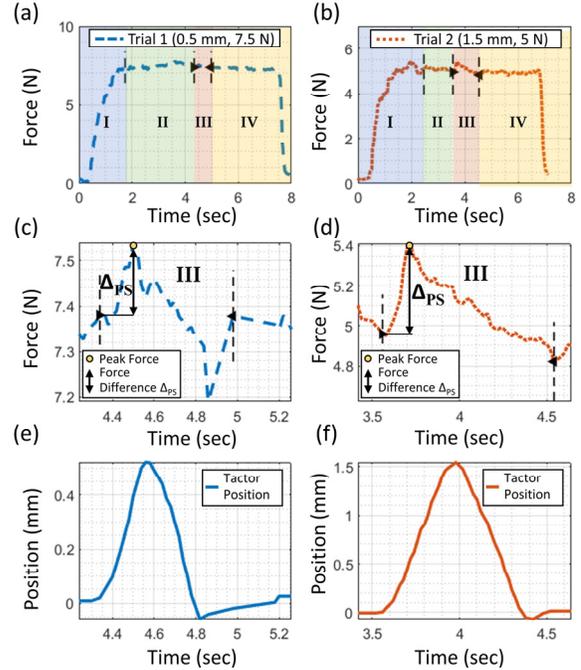

Fig. 8. Example of two trials of subject 5. The two trials (left and right columns) have target forces of 5 N and 7.5 N and tactor displacements of 1.5 mm and 0.5 mm respectively. (a) and (b): The force along all phases of the trial: I - reaching the target force, II – keeping a stable grip within the target force limits, III – tactor movement, also marked with two triangles and IV – waiting time after stimulus ends and the releasing of grip. (c) and (d): An enlarged view of the grip force during tactor movement for both trials (left and right). The force difference between the peak force and the force when tactor movement starts is marked with a double arrow. (e) and (f): Recorded tactor position during the two trials.

To analyze the difference in grip force applied from each side of the gripper, we used (8) and (9) to calculate the forces applied on each side of the gripper for all subjects. We were particularly interested in the difference during the stable grip phase of the trials, marked III in Fig. 8(a). Since the stable grip phase is independent of the preceding skin-stretch stimulus, we averaged the recorded grip forces for each subject across all tactor displacement conditions and for each target force condition separately. The stable grip phase of the trials lasted between 1 to 4 seconds, distributed randomly, and predetermined. Therefore, we averaged the grip forces over 1 second before the beginning of the stimulus. We compared this average force of applied by the finger with the one applied by the thumb.

## C. Results

Fig. 8(a1) and (a2) show an example of the recorded force throughout the course of two trials of a single participant. An enlarged view of the recorded force during tactor movement, presented in Fig. 8(b), shows that in both trials, there was a sudden increase in grip force starting from about 100 ms after initial tactor movement. Fig. 8(c) presents the tactor movement trajectory recorded during both trials. We observed a similar increase in grip force of different sizes in all other participants. Averages of grip force across participants for all 6 conditions are shown in Fig. 9. Participants applied forces closer to the target force for the 5 N compared with the 7.5 N. Additionally, the increase in target force is larger for the 5 N target force compared with the 7.5 N target force.

The $\Delta_{PS}$ data for all subjects and all 6 conditions are shown in Fig. 10, with average across participants and standard errors. Fig. 10 shows that for all three tactor displacements, the average $\Delta_{PS}$ is higher for the lower target force 5 N compared to the higher force 7.5 N. It also shows that for both target forces, $\Delta_{PS}$ is larger for a 1 mm tactor displacement compared to 0.5 mm. As for the average $\Delta_{PS}$ for a 1.5 mm tactor displacement compared to 1 mm tactor displacement, there is a large difference for the 5 N target force but nearly no difference for the 7.5 N target force.

These observations were corroborated by the statistical analysis. The two-way repeated measures ANOVA showed a statistically significant effect of the level of target force on $\Delta_{PS}$ ($F(1,9)=14.67$, $p=0.004$), a statistically significant effect of the tactor displacement level on $\Delta_{PS}$ ($F(2,18)=27.31$, $p<0.001$) and a statistically significant interaction between the effects of target force and tactor displacement on $\Delta_{PS}$ ($F(2,18)=7.95$, $p=0.003$).

The comparisons for the 5 N target force showed a statistically significant difference in the effect of 1 mm tactor displacement on $\Delta_{PS}$ compared to the effect of 0.5 mm ($\Delta=0.044$ N, $t(35.93)=2.74$, $p_H=0.038$), and a statistically significant difference in the effect of 1.5 mm tactor displacement $\Delta_{PS}$ compared to the effect of 1 mm tactor displacement ($\Delta=0.080$ N, $t(35.93)=4.93$, $p_H<0.001$). For the 5 N target force, there was a significant difference in the effect of 1 mm tactor displacement on $\Delta_{PS}$ compared to the effect of 0.5 mm ($\Delta=0.039$ N, $t(35.93)=2.39$, $p_H=0.044$). There was a non-significant difference in the effect of 1.5 mm tactor displacement on $\Delta_{PS}$ compared to the effect of 1 mm tactor displacement ($\Delta=0.001$ N, $t(35.93)=0.09$, $p_H=0.930$).

An example of grip forces averaged across all tactor displacement conditions is shown in Fig. 11. The stable grip period is marked in Fig. 11(a) with a black dashed rectangle. Fig. 11(b) shows that on average subjects applied larger forces with their finger compared to their thumb consistently for both target forces. Two subjects applied larger forces with their thumb.

## V. DISCUSSION

### A. Design, Calibration & Validation

We developed a new grip force measurement concept that allows for embedding tactile stimulation mechanisms in the gripper. This force measurement concept can be used to accurately measure each of the grip forces applied on both sides of the gripper using a single force sensor while eliminating artifacts caused by tactor movements. It allows for measuring the forces that are applied on the aperture and the tactor, and therefore, provides a more accurate alternative to the concept used in [5],[6],[13],[15], where only the force applied on the aperture is measured, leading to a loss of force in the measurement. An actuation mechanism can be integrated into the force measurement device. Here, we designed and used a 2-DoF skin-stretch actuation mechanism, however, we specifically designed this concept such that similar designs can easily have other actuation mechanisms for tactile stimulation incorporated into them.

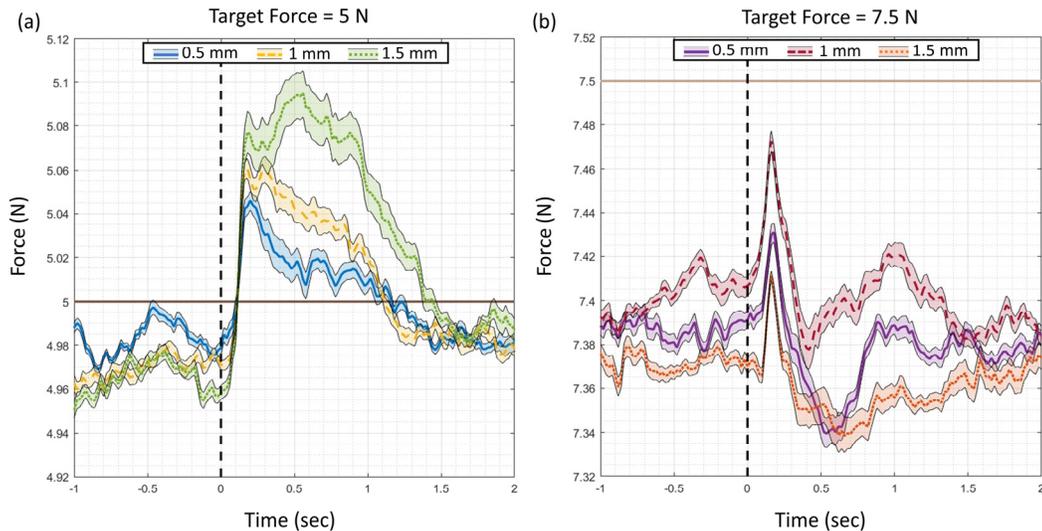

Fig. 9. Mean force and standard error for all 10 subjects with (a) 5 N target force and (b) 7.5 N target force. Colors represent different tactor displacement conditions. The dashed vertical line at t=0 sec represents the beginning of tactor movement. The horizontal lines represent the center of the target force range.

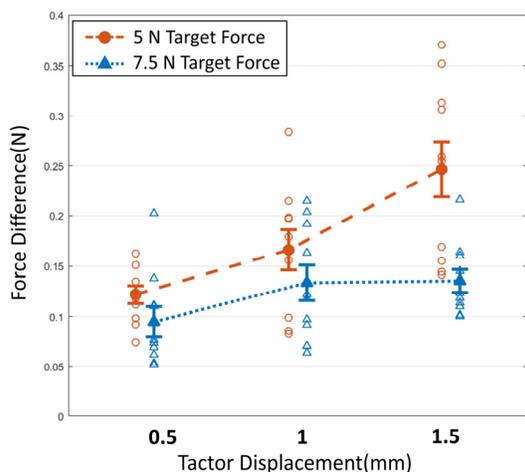

Fig. 10. ΔPS for different target forces and different tactor displacements. Individual results are shown with empty markers while averages are represented by filled markers and SE by bars.

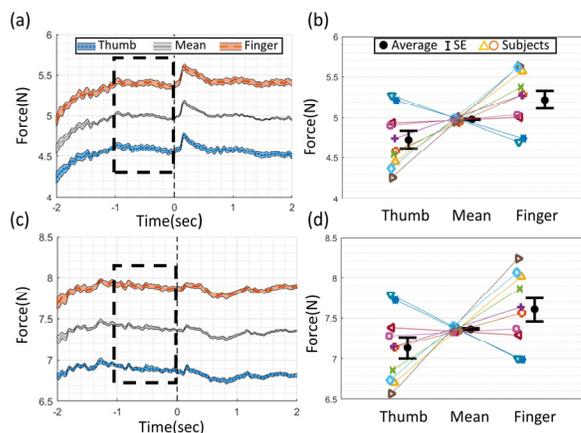

Fig. 11. (a) and (c): Example of the mean response of Subject 5 with standard error, for a target grip force of 5 N (a) and 7.5 N (c). (b) Average of force along 1 second before stimulation was applied (marked with a black dashed line on the left) for a target grip force of 5 N (b1) and 7.5 N (b2). Black dot markers with error bars show average over subjects with standard error.

Results of calibration and validation tests show small errors during both static and dynamic measurements. Following the calibration process, errors caused by manufacturing tolerances or other systematic errors were reduced. Our device was built mainly using FDM 3D printing, and therefore, more accurate manufacturing of the device, and in particular the levers and the back base, could further reduce measurement errors.

Results of our validation process show minimal artifacts in force measurements caused by tactor movement (maximum of 0.65% and 1% for the $x$ and $y$ tactor movement directions respectively). We administered these tests in conditions that according to our force analysis are likely to result in larger artifacts that expected in user studies. For example, rubber on rubber interface has a larger friction coefficient than finger on rubber [61],[62]. In addition, we used the maximum range movement of the tactor a large grip force, resulting in large tangential forces that may cause artifacts. Therefore, we expect that the actual artifacts in user studies will be smaller.

This validation process was tailored to our device, but can be adjusted to match particular characteristics of a similarly designed measurement device. Such characteristics include the user interface, the integrated actuation mechanism, or the expected range of forces to be measured. Farajian et al. [15], also tested the artifacts in the force measurement of their device during tactor movement. In their test, they used a padded clamp to maintain a stable grip force while displacing the tactor from the center of the aperture outwards by different amounts. The artifact they observed in the measurement had a different pattern than ours: they observed a decrease in force followed by an increase, whereas here we observed only an increase. We hypothesize that the reason for this difference stems from the loss of force that is applied on the tactor in their measurement method. Additionally, the padded clamp could have caused an increased force when the tactor pushed against the clamps while moving, causing it to extend and apply a force that depends on the stiffness of the clamps. In our test, we measured the applied external force during tactor movement such that we could eliminate changes originating in the stiffness of the calibration system by taking the relative error $A_{\text{TM}}$.

The device allows for separately measuring forces from both sides of the grip during the application of static or dynamic grip forces. Results showed low errors between dynamic forces measured using external force sensors and the ones measured by the device (2.67%, 2.93%, and 3.71%, for the mean, left and right forces, respectively). To the best of our knowledge, none of the tactile stimulation devices that incorporated grip force measurement could isolate the force applied by each of the fingers [5],[6],[13],[15].

Some other devices that measure the grip force but do not include tactile stimulation, use sensors incorporated into the gripper between the fingers [38]–[42],[44] or attached to an object [18],[19]. Indeed, using two sensors to directly measure the forces might provide better accuracy in the measurement. However, it makes the measurement system more expensive and elicits limitations on the design of the gripper. For example, it requires having a very wide grip aperture. Moreover, the accuracy of our device can measure the sizes of effect seen in previous studies, ranging between 10 % [5],[15], and hundreds of percent of the initial grip force [23]. Therefore, our measurement concept can be used in studying independent control of human fingertip forces [47], abnormalities in the control of grip such as in cases of some pathological conditions [48]–[50], and their relation with tactile sensation.

### B. User Study

We used our novel device to measure the grip force that users applied during a static force maintenance task. In most of the trials, the grip force increased subsequent to tactor movement. The average pattern of the change in grip force started with an increase in grip force as the tactor started moving, followed by a decrease back to around initial force level. In some of the trials, in the large target force condition, the increase was followed by a decrease below the initial force level, but this decrease was not consistent across trials. This pattern is different from the one reported in Farajian et al.[15], where the response typically started with a substantial decrease in grip force prior to a smaller increase. This difference suggests that consistently with our hypothesis and with the concerns raised in Farajian et al. [15], at least part of the decrease in grip force that they observed can be explained by measurement artifacts. This could be a result of their measurement that captured only the part of the grip force that was applied on the aperture.

One possible explanation to the increase in grip force caused by tactor movement is that it was interpreted as a sudden load force or slippage of a held object. It was shown that feedback from tactile mechanoreceptors in the finger pads has a role in the response to a sudden change in friction conditions or an unpredictable load force [25],[29],[32],[34],[35],[63].

In our static force maintenance task, the tactor moved against the finger pad skin of the participants while stretching it and applying a local shear force to it. This stretch of the skin is present in natural interaction with a held object and when applied artificially concurrently with a load force it is known to augment this force [5],[15]. In addition to stretching the skin, the tactor may also be fully or partially slipping over the contact area [64],[65]. Thus, the observed increase in grip force could be a response induced by feedback from mechanoreceptors that is meant to prevent the loss of grip or stop an on-going slippage.

Another possible explanation is that the response is purely a dynamic response that depends on the mechanical properties of the finger pad skin and the dynamics of the hand of the participant, and had nothing to do with active feedback from the sensorimotor system. As the tactor moves against the curvature of the finger pad, the distribution of grip force between the tactor ($F_T$) and aperture ($F_A$) can change due to normal deformation of the skin [23],[64]–[66]. It is possible that the dynamics of this change can induce a temporary altering in the grip force. This could explain the change in the measured force that Farajian et al. [15] observed when using a padded clamp to apply a constant force to their device while moving the tactor, although that change may have been a result of measuring only $F_A$. We observed a change in grip force when applying a constant force to our device only when the tactor moved in the $y$ direction, which does not comply with this hypothesis. However, further modelling and testing is required to estimate the magnitude of this effect.

The effect of tactile stimuli on the increase of the grip force was larger for the smaller target force. A larger grip force causes a larger normal force between the tactor and finger pad, resulting in a larger friction force, and possibly, more stimulation of mechanoreceptors [7],[44],[45]. Therefore, we hypothesized that applying a larger target force will result in a larger increase in grip force, however, our results contradict this hypothesis. There are several possible explanations to this result. A ceiling effect in grip force that participants could apply may have caused smaller response in the larger target grip force. This is however not likely, because humans can easily apply larger grip forces, up to 65 N [67]. Another explanation could be that a larger normal force caused both normal and tangential deformation of the finger pad skin, affecting its mechanical properties and as a result, its response to tactor movement [23],[64]–[66].

Another likely explanation is related to safety margins in grip force control [24]–[26],[29],[32],[36]. When we hold an object, we apply a grip force that is only slightly above the force needed to prevent slippage in accordance with the internal representation of the dynamics of the held object [8],[10],[11],[39],[40],[48],[49]. In our experiments, applying a larger target force to the same object may have increased the safety margin above the necessary grip force to prevent slippage of the held object. As a result, the confidence of the participants that the object cannot slip from their grip may have increased, and therefore, the reaction to tactor movement was smaller. It is possible that the effect of increasing target force on the change in grip force during a skin stretch stimulus is a combination of the increase in friction force with the tactor and an increase in the safety margin, which have opposite contributions. To further test these hypotheses, more experiments with additional target force conditions and in more natural object manipulation scenarios are needed.

Larger tactor displacements caused larger changes in grip force. Nonetheless, for the 7.5 N target force there was no significant difference between the change in grip force caused by a 1.5 mm tactor displacement compared with the one caused by a 1 mm displacement. It is possible that for the larger target force, a lower tactor displacement is needed to reach saturation of the tactile sensation in the finger pad, because the mechanoreceptors can already be saturated [33]. Another possible reason for such saturation could be that for larger grip forces the skin is less compliant [23],[64]–[66], and therefore, more difficult to stretch. Gleeson et al. [17] tested the effect of tactor displacement in a direction identification test and discovered that for speeds ranging from 0.5 mm/s to 4 mm/s, no significant improvement in performance was seen for tactor displacements over 0.2 mm, which might also be explained by saturation of tactile sensation or compliance of the skin to stretch. To test this in our experimental setting, future studies are needed with additional larger tactor displacements and higher target forces. Furthermore, higher resolution of tactor displacement could provide a more accurate result.

Finally, our results provided validation for the ability of the device to measure forces from both sides of the gripper. Most participants applied a larger force with their finger compared to their thumb during the stable grip phase of the experiments. It is worth reminding that during our experiments, skin stretch was applied only to the finger. However, the stable grip phase in a trial occurs before tactor movement. Therefore, those results do not indicate a connection between the stimulated finger to the applied force. One possible explanation for this result is that the participants sat to the left of the device such that their posture caused them to apply a pulling force towards the left, resulting in a higher force applied by the finger of their right hand. However, extensive studies with additional postures and possibly also with the left hand are needed to ascertain such explanation.

## VI. CONCLUSION AND FUTURE WORK

We designed a novel grip force measurement concept and implemented it in our newly designed device. To demonstrate the integration of a tactile mechanism into the device's gripper, we designed a 2-DoF actuation mechanism for skin stretch stimulation. Calibration and validation processes showed that the device can accurately measure grip forces while applying skin stretch to the user's finger pad.

To test the device in a user study, we conducted an experiment in which participants had to maintain a constant target grip force while skin stretch stimuli of different displacement levels was applied to their finger pad. Results showed that users increased their grip force when the skin stretch stimulus was applied, that the effect was larger for higher tactor displacements, and that saturation of the effect can be reached above a certain level of skin stretch. Furthermore, effects were larger for the lower target force, likely due to lower level of confidence in the grip.

Our grip force measurement concept provides a more accurate alternative to the existing measurement solutions. The results of both validation process and a user study show that it can be used to investigate human motor response to tactile stimulation. Such response might be an important consideration in the design of manipulation mechanisms and controllers for teleoperation or virtual reality with haptics. Therefore, using our new concept to better characterize this response can contribute to better designs of such systems.

ACKNOWLEDGMENT

The authors would like to thank Amit Milstein for his help with the implementation of the code for the user study experiment.

REFERENCES

[1] R. P. Khurshid, N. T. Fitter, E. A. Fedalei, and K. J. Kuchenbecker, "Effects of grip-force, contact, and acceleration feedback on a teleoperated pick-and-place task," *IEEE Trans. Haptics*, vol. 10, no. 1, pp. 40–53, 2017.

[2] T. L. Gibo, A. J. Bastian, and A. M. Okamura, "Grip force control during virtual objectinteraction: Effect of force feedback, Accuracy Demands, and Training," *IEEE Trans. Haptics*, vol. 7, no. 1, pp. 37–47, 2014.

[3] J. M. Romano, K. Hsiao, G. Niemeyer, S. Chitta, and K. J. Kuchenbecker, "Human-inspired robotic grasp control with tactile sensing," *IEEE Trans. Robot.*, vol. 27, no. 6, pp. 1067–1079, 2011.

[4] Z. F. Quek et al., "Augmentation Of Stiffness Perception With a 1-Degree-of-Freedom Skin Stretch Device Zhan," *IEEE Trans. HUMAN-MACHINE Syst.*, vol. 44, no. 6, pp. 731–742, 2014.

[5] Z. Fan Quek, S. B. Schorr, I. Nisky, W. R. Provancher, and A. M. Okamura, "Sensory substitution and augmentation using 3-degree-of-freedom skin deformation feedback," *IEEE Trans. Haptics*, vol. 8, no. 2, pp. 209–221, 2015.

[6] Z. F. Quek, S. B. Schorr, I. Nisky, W. R. Provancher, and A. M. Okamura, "Sensory substitution of force and torque using 6-DoF tangential and normal skin deformation feedback," *Proc. - IEEE Int. Conf. Robot. Autom.*, vol. 2015-June, no. June, pp. 264–271, 2015.

[7] C. Pacchierotti, L. Meli, F. Chinello, M. Malvezzi, and D. Prattichizzo, "Cutaneous haptic feedback to ensure the stability of robotic teleoperation systems," *Int. J. Rob. Res.*, vol. 34, no. 14, pp. 1773–1787, 2015.

[8] D. Leonardis, M. Solazzi, I. Bortone, and A. Frisoli, "A wearable fingertip haptic device with 3 DoF asymmetric 3-RSR kinematics," *IEEE World Haptics Conf. WHC 2015*, pp. 388–393, 2015.

[9] A. Guzererler, W. R. Provancher, and C. Basdogan, "Perception of Skin Stretch Applied to Palm: Effects of Speed and Displacement," in *Haptics: Perception, Devices, Control, and Applications*, 2016, pp. 180–189.

[10] H. Culbertson, S. B. Schorr, and A. M. Okamura, "Haptics: The Present and Future of Artificial Touch Sensation," *Annu. Rev. Control. Robot. Auton. Syst.*, vol. 1, no. 1, pp. 385–409, 2018.

[11] M. Aggravi, F. Pause, P. R. Giordano, and C. Pacchierotti, "Design and Evaluation of a Wearable Haptic Device for Skin Stretch, Pressure, and Vibrotactile Stimuli," *IEEE Robot. Autom. Lett.*, vol. 3, no. 3, pp. 2166–2173, 2018.

[12] N. Colella, M. Bianchi, G. Grioli, A. Bicchi, and M. G. Catalano, "A novel skin-stretch haptic device for intuitive control of robotic prostheses and avatars," *IEEE Robot. Autom. Lett.*, vol. 4, no. 2, pp. 1572–1579, 2019.

[13] C. Avraham and I. Nisky, "The effect of tactile augmentation on manipulation and grip force control during force-field adaptation," *J. Neuroeng. Rehabil.*, pp. 1–19, 2020.

[14] W. R. Provancher and N. D. Sylvester, "Fingerpad Skin stretch increases the perception of virtual Friction," *IEEE Trans. Haptics*, vol. 2, no. 4, pp. 212–223, 2009.

[15] M. Farajian, R. Leib, H. Kossowsky, T. Zaidenberg, F. A. Mussa-Ivaldi, and I. Nisky, "Stretching the skin immediately enhances perceived stiffness and gradually enhances the predictive control of grip force," *Elife*, vol. 9, pp. 1–38, 2020.

[16] B. Gleeson, S. Horschel, and W. Provancher, "Design of a fingertip-mounted iactile display with iangential skin displacement feedback," *IEEE Trans. Haptics*, vol. 3, no. 4, pp. 297–301, 2010.

[17] B. T. Gleeson, S. K. Horschel, and W. R. Provancher, "Perception of direction for applied tangential skin displacement: Effects of speed, displacement, and repetition," *IEEE Trans. Haptics*, vol. 3, no. 3, pp. 177–188, 2010.

[18] M. Solazzi, A. Frisoli, and M. Bergamasco, "Design of a Novel Finger Haptic Interface for Contact and Orientation Display," *2010 IEEE Haptics Symp.*, pp. 129–132, 2010.

[19] B. T. Gleeson, C. A. Stewart, and W. R. Provancher, "Improved tactile shear feedback: Tactor design and an aperture-based restraint," *IEEE Trans. Haptics*, vol. 4, no. 4, pp. 253–262, 2011.

[20] A. L. Guinan, N. C. Hornbaker, M. N. Montandon, A. J. Doxon, and W. R. Provancher, "Back-to-Back Skin Stretch Feedback for Communicating Five Degree- of-Freedom Direction Cues," *World Haptics Conf.*, pp. 13–18, 2013.

[21] D. Prattichizzo, F. Chinello, C. Pacchierotti, and M. Malvezzi, "Towards wearability in fingertip haptics: A 3-DoF wearable device for cutaneous force feedback," *IEEE Trans. Haptics*, vol. 6, no. 4, pp. 506–516, 2013.

[22] S. B. Schorr, Z. F. Quek, R. Y. Romano, I. Nisky, W. R. Provancher, and A. M. Okamura, "Sensory substitution via cutaneous skin stretch feedback," *Proc. - IEEE Int. Conf. Robot. Autom.*, pp. 2341–2346, 2013.

[23] R. S. Johansson and J. R. Flanagan, "Coding and use of tactile signals from the fingertips in object manipulation tasks," *Nat. Rev. Neurosci.*, vol. 10, no. 5, pp. 345–359, 2009.

[24] J. R. Flanagan and A. M. Wing, "The role of internal models in motion planning and control: Evidence from grip force adjustments during movements of hand-held loads," *J. Neurosci.*, vol. 17, no. 4, pp. 1519–1528, 1997.

[25] R. S. Johansson and G. Westling, "Programmed and triggered actions to rapid load changes during precision grip," *Exp. Brain Res.*, pp. 72–86, 1988.

[26] D. A. Nowak, J. Hermsdorfer, S. Glasauer, J. Philipp, L. Meyer, and N. Mai, "The effects of digital anaesthesia on predictive grip force adjustments during vertical movements of a grasped object," *Eur. J. Neurosci.*, vol. 14, pp. 756–762, 2001.

[27] D. A. Nowak and J. Hermsdörfer, "Selective deficits of grip force control during object manipulation in patients with reduced sensibility of the grasping digits," *Neurosci. Res.*, vol. 47, no. 1,

pp. 65–72, 2003.
[28] D. A. Nowak, J. Hermsdörfer, C. Marquardt, and H. Topka, "Moving objects with clumsy fingers: How predictive is grip force control in patients with impaired manual sensibility?," *Clin. Neurophysiol.*, vol. 114, no. 3, pp. 472–487, 2003.
[29] R. S. Johansson and G. Westling, "Roles of glabrous skin receptors and sensorimotor memory in automatic control of precision grip when lifting rougher or more slippery objects," *Exp. Brain Res.*, pp. 550–564, 1984.
[30] G. Westling and R. S. Johansson, "Factors influencing the force control during precision grip," *Exp. Brain Res.*, vol. 53, no. 2, pp. 277–284, 1984.
[31] Kandel, *Principles of Neural Science*. New-York: McGraw-hill, 2000.
[32] R. S. Johansson and G. Westling, "Signals in tactile afferents from the fingers eliciting adaptive motor responses during precision grip," *Exp. Brain Res.*, vol. 66, no. 1, pp. 141–154, 1987.
[33] G. Westling and R. S. Johansson, "Responses in glabrous skin mechanoreceptors during precision grip in humans," *Exp. Brain Res.*, pp. 128–140, 1987.
[34] R. S. Johansson, R. Riso, C. Häger, and L. Bäckström, "Somatosensory control of precision grip during unpredictable pulling loads - I. Changes in load force amplitude," *Exp. Brain Res.*, vol. 89, no. 1, pp. 181–191, 1992.
[35] R. S. Johansson, C. Häger, and L. Bäckström, "Somatosensory control of precision grip during unpredictable pulling loads - III. Impairments during digital anesthesia," *Exp. Brain Res.*, vol. 89, no. 1, pp. 204–213, 1992.
[36] A. S. Augurelle, A. M. Smith, T. Lejeune, and J. L. Thonnard, "Importance of cutaneous feedback in maintaining a secure grip during manipulation of hand-held objects," *J. Neurophysiol.*, vol. 89, no. 2, pp. 665–671, 2003.
[37] M. Farajian, R. Leib, T. Zaidenberg, F. Mussa-Ivaldi, and I. Nisky, "Stretching the skin of the fingertip creates a perceptual and motor illusion of touching a harder spring," *bioRxiv*, p. 203604, 2018.
[38] R. Reilmann and A. M. Gordon, "Initiation and development of fingertip forces during whole-hand grasping," *Exp. Brain Res.*, pp. 443–452, 2001.
[39] S. A. Winges, S. E. Eonta, J. F. Soechting, and M. Flanders, "Effects of Object Compliance on Three-Digit Grasping," *J. Neurophysiol.*, pp. 2447–2458, 2009.
[40] Q. Fu, W. Zhang, and M. Santello, "Anticipatory Planning and Control of Grasp Positions and Forces for Dexterous Two-Digit Manipulation," *J. Neurosci.*, vol. 30, no. 27, pp. 9117–9126, 2010.
[41] J. A. Johnston, L. R. Bobich, and M. Santello, "Neuroscience Letters Coordination of intrinsic and extrinsic hand muscle activity as a function of wrist joint angle during two-digit grasping," *Neurosci. Lett.*, vol. 474, pp. 104–108, 2010.
[42] J. Metzger, O. Lambercy, D. Chapuis, and R. Gassert, "Design and Characterization of the ReHapticKnob , a Robot for Assessment and Therapy of Hand Function," *IEEE/RSJ Int. Conf. Intell. Robot. Syst.*, pp. 3074–3080, 2011.
[43] X. A. Naceri, X. A. Moscatelli, R. Haschke, H. Ritter, M. Santello, and M. O. Ernst, "Multidigit force control during unconstrained grasping in response to object perturbations," *J. Neurophysiol.*, pp. 2025–2036, 2017.
[44] X. D. Shibata and M. Santello, "Role of digit placement control in sensorimotor transformations for dexterous manipulation," *J. Neurophysiol.*, pp. 2935–2943, 2017.
[45] S. Toma, D. Shibata, F. Chinello, D. Prattichizzo, and M. Santello, "Linear Integration of Tactile and Non-tactile Inputs Mediates Estimation of Fingertip Relative Position," *Front. Neurosci.*, vol. 13, no. February, 2019.
[46] W. Zhang, A. M. Gordon, Q. Fu, and M. Santello, "Manipulation after object rotation reveals independent sensorimotor memory representations of digit positions and forces," *J. Neurophysiol.*, vol. 103, no. 6, pp. 2953–2964, 2010.
[47] B. Y. B. B. Edin, G. Westling, and R. S. Johansson, "Independent Control of Human Finger-Tip Forces at Indicidual Digits During Precision Lifting," *J. Physiol.*, pp. 547–564, 1992.
[48] P. E. Ingvarsson, A. M. Gordon, and H. Forssberg, "Coordination of manipulative forces in Parkinson's disease," *Exp. Neurol.*, vol. 145, no. 2 I, pp. 489–501, 1997.
[49] M. P. Rearick, G. E. Stelmach, B. Leis, and M. Santello, "Coordination and Control of Forces during Multifingered Grasping in Parkinson ' s Disease," vol. 442, pp. 428–442, 2002.
[50] J. Hermsdörfer, E. Hagl, D. A. Nowak, and C. Marquardt, "Grip force control during object manipulation in cerebral stroke," *Clin. Neurophysiol.*, vol. 114, no. 5, pp. 915–929, 2003.
[51] T. H. Massie and J. K. Salisbury, "The PHANTOM Haptic Interface : A Device for Probing Virtual Objects Threee Enabling Observations component of our ability to ' visualize ,' remember and establish cognitive models of the physical structure of our environment stems from haptic interactions," *ASME Winter Annu. Meet. Symp. Haptic Interfaces Virtual Environ. Teleoperator Syst.*, pp. 1–6, 1994.
[52] J. E. Colgate and J. M. Brown, "Factors affecting the Z-width of a haptic display," *Proc. - IEEE Int. Conf. Robot. Autom.*, no. pt 4, pp. 3205–3210, 1994.
[53] N. Colonnese and A. M. Okamura, "M-Width: Stability, noise characterization, and accuracy of rendering virtual mass," *Int. J. Rob. Res.*, vol. 34, no. 6, pp. 781–798, 2015.
[54] S. Paneels, M. Anastassova, S. Strachan, S. P. Van, S. Sivacoumarane, and C. Bolzmacher, "What's around me Multi-actuator haptic feedback on the wrist," *2013 World Haptics Conf. WHC 2013*, pp. 407–412, 2013.
[55] L. B. Stephen Brewster, "Tactons: Structured Vibrotactile Messages for Non-Visual Information Display," *5th Australas. User Interface Conf.*, p. 228, 2007.
[56] E. Battaglia, J. P. Clark, M. Bianchi, M. G. Catalano, A. Bicchi, and M. K. O'Malley, "The Rice Haptic Rocker: Skin stretch haptic feedback with the Pisa/IIT SoftHand," *2017 IEEE World Haptics Conf. WHC 2017*, no. June, pp. 7–12, 2017.
[57] V. E. Abraira and D. D. Ginty, "The sensory neurons of touch," *Neuron*, vol. 79, no. 4, pp. 618–639, 2013.
[58] K. J. Kuchenbecker, M. Kutzer, D. Ferguson, M. Moses, and A. M. Okamura, "The touch thimble: Providing fingertip contact feedback during point-force haptic interaction," *Symp. Haptics Interfaces Virtual Environ. Teleoperator Syst. 2008 - Proceedings, Haptics*, pp. 239–246, 2008.
[59] R. G. (Richard G. Budynas, *Shigley's mechanical engineering design*, 8th ed. in. Singapore: McGraw-Hill, 2008.
[60] S. A.Glantz and B. K.Slinker, "Multiple Comparisons in Two-Way Repeated Measures Designs," in *Primer on Applied Regression and Analysis of Variance*, Second Edi., McGraw-Hill, 2001, pp. 494–501.
[61] B. N. . Persson, *Sliding Friction: Physical Principles and Applications*, Second Edi. Springer, 2000.
[62] M. J. Adams *et al.*, "Finger pad friction and its role in grip and touch," *J. R. Soc. Interface*, vol. 10, no. 80, 2013.
[63] A. C. Eliasson, H. Forssberg, K. Ikuta, I. Apel, G. Westling, and R. Johansson, "Development of human precision grip - V. Anticipatory and triggered grip actions during sudden loading," *Exp. Brain Res.*, vol. 106, no. 3, pp. 425–433, 1995.
[64] J. Thonnard, B. Delhaye, and P. Lefe, "Dynamics of fingertip contact during the onset of tangential slip," *J. R. Soc. Interface*, 2014.
[65] B. Delhaye, A. Barrea, B. B. Edin, and P. Lefe, "Surface strain measurements of fingertip skin under shearing," *J. R. Soc. Interface*, 2018.
[66] N. Nakazawa, R. Ikeura, and H. Inooka, "Characteristics of human fingertips in the shearing direction," *Biol. Cybern.*, vol. 214, pp. 207–214, 2000.
[67] V. A. Moerchen, J. C. Lazarus, and K. G. Gruben, "Task-dependent organization of pinch grip forces," *Exp. Brain Res.*, pp. 367–376, 2007.